\documentclass[conference]{IEEEtran}
\IEEEoverridecommandlockouts

\usepackage{cite}
\usepackage{amsmath,amssymb,amsfonts}
\usepackage{algorithmic}
\usepackage{graphicx}
\usepackage{textcomp}
\usepackage{xcolor}
\usepackage{tabularray}

\usepackage{amssymb}
\usepackage[caption=false,font=footnotesize]{subfig}

\usepackage[pagebackref=false,breaklinks=true,bookmarks=false]{hyperref}

\def\BibTeX{{\rm B\kern-.05em{\sc i\kern-.025em b}\kern-.08em
    T\kern-.1667em\lower.7ex\hbox{E}\kern-.125emX}}
\begin{document}

\title{A Comprehensive Survey on Deep Learning-Based LiDAR Super-Resolution for Autonomous Driving\\
}
\author{June Moh Goo$^{*}$, Zichao Zeng and Jan Boehm
\thanks{*\;Corresponding Author (e-mail:\href{mailto:june.goo.21@ucl.ac.uk}{june.goo.21@ucl.ac.uk})}%
\thanks{This work was supported by the Engineering and Physical Sciences Research Council through an industrial CASE studentship with Ordnance Survey (Grant number EP/X524840/1 and EP/W522077/1).}
\thanks{The authors are with Department of Civil, Environmental and Geomatic Engineering,
        University College London, WC1E 6BT London, U.K.
        {june.goo.21@ucl.ac.uk; zichao.zeng.21@ucl.ac.uk; j.boehm@ucl.ac.uk}}%
}

\maketitle

\begin{abstract}
LiDAR sensors are often considered essential for autonomous driving, but high-resolution sensors remain expensive while affordable low-resolution sensors produce sparse point clouds that miss critical details. LiDAR super-resolution addresses this challenge by using deep learning to enhance sparse point clouds, bridging the gap between different sensor types and enabling cross-sensor compatibility in real-world deployments. This paper presents the first comprehensive survey of LiDAR super-resolution methods for autonomous driving. Despite the importance of practical deployment, no systematic review has been conducted until now. We organize existing approaches into four categories: CNN-based architectures, model-based deep unrolling, implicit representation methods, and Transformer and Mamba-based approaches. We establish fundamental concepts including data representations, problem formulation, benchmark datasets and evaluation metrics. Current trends include the adoption of range image representation for efficient processing, extreme model compression and the development of resolution-flexible architectures. Recent research prioritizes real-time inference and cross-sensor generalization for practical deployment. We conclude by identifying open challenges and future research directions for advancing LiDAR super-resolution technology.
\end{abstract}

\begin{IEEEkeywords}
LiDAR super-resolution, autonomous driving, cross-sensor gap
\end{IEEEkeywords}

\section{Introduction}
LiDAR sensors are essential for autonomous driving. They provide accurate 3D information about the environment. We limit our discussion to the most common category of automotive LiDAR sensors using rotating multiple laser beams positioned to maximize the horizontal field-of-view for obstacle avoidance and navigation. High-resolution LiDAR sensors with 64 or 128 beams are significantly more expensive than lower resolution sensors \cite{ortiz2019initial}. Since the price limits their use in consumer vehicles, most manufacturers choose cheaper 16 or 32-beam sensors instead \cite{shang2025investigating}. These low-resolution sensors produce sparse point clouds that provide fewer returns on crucial details for safe navigation.

\begin{figure}[t]
    \centering
    \includegraphics[width=\linewidth]{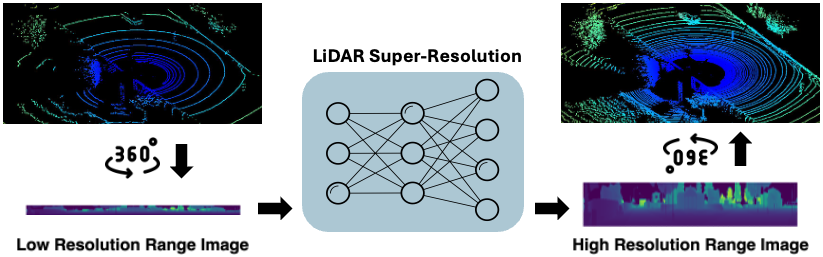}
    \caption{General pipeline of projection-based LiDAR super-resolution for autonomous driving. The process begins with a sparse 3D point cloud from a low-resolution LiDAR sensor (left), which is projected into a 2D range image representation preserving the 360-degree horizontal field of view. The low-resolution range image is then processed through a deep learning network to predict a high-resolution range image with increased vertical resolution. Finally, the enhanced range image is back-projected to 3D space, generating a dense point cloud comparable to expensive high-resolution sensors.}
    \label{fig:intro}
\end{figure}

LiDAR super-resolution (SR) offers a solution to this problem. It uses deep learning to increase the density of sparse point clouds. The goal is to make cheap sensors perform like expensive ones. This technology could enable widespread deployment of autonomous vehicles by reducing sensor costs while maintaining safety standards \cite{goo2025flash}. Figure~\ref{fig:intro} demonstrates this capability which shows how most modern super-resolution methods can transform sparse (low-resolution LiDAR) into dense (high-resolution LiDAR) quality point clouds.

LiDAR super-resolution faces challenges compared to image super-resolution. First, most methods define the task specifically to target vertical resolution enhancement while preserving horizontal resolution. Second, LiDAR typically used in automotive has a 360-degree horizontal field of view. Third, LiDAR shows sharp depth changes at object boundaries. Cars, buildings, and pedestrians create sudden jumps in depth values. Fourth, autonomous driving needs real-time processing. Models must run at least 25 fps to match sensor frame rates \cite{dai2022requirements}. Fifth, point clouds are sparse and irregular. Unlike images with uniform pixel grids, LiDAR points scatter unevenly in 3D space. Sixth, downstream tasks of LiDAR-based models suffer from severe resolution-dependent domain gaps. While image-domain model \cite{dosovitskiy2021an} trained on high-resolution data typically maintain reasonable performance on lower resolutions, 3D object detection models exhibit significant degradation \cite{fang2024lidar}. 

Most LiDAR super-resolution methods adapted 2D image super-resolution techniques to range images which treated LiDAR data as depth images. While it is simplistic, they ignored the geometric properties of 3D point clouds \cite{shan2020simulation}. Traditional geometric methods \cite{tian2022lidar} increase point cloud density through interpolation and geometric analysis without training from the data. In this survey, we focus on deep learning-based super-resolution methods that learn to reconstruct high-resolution patterns from low-resolution inputs. 

Recent methods incorporate domain knowledge about LiDAR sensors. They handle circular padding for 360-degree views \cite{yang2024tulip, chen2021channel}, use polar coordinates to reduce errors \cite{eskandar2022hals}, and preserve 3D structures \cite{park2023implicit,chen2025srmambav2}. Model-based approaches use the physical sensor model to guide learning \cite{gkillas2025guided, gkillas2023efficient}. Implicit methods learn continuous functions that work at any resolution \cite{kwon2022implicit}. Transformer architectures capture long-range dependencies from range images \cite{yang2024tulip,goo2025flash}. Mamba architectures leverage selective state-space modeling to efficiently capture both local and global context \cite{chen2025srmamba,chen2025srmambav2}.

This paper provides the first comprehensive survey of deep learning-based LiDAR super-resolution for autonomous driving. We organize existing methods into four categories:

\begin{itemize}
    \item CNN-based architectures from early to advanced designs (Section~\ref{sec:CNN})
    \item Model-based deep unrolling and federated learning methods (Section~\ref{sec:DU_FM})
    \item Implicit representation methods (Section~\ref{sec:implicit})
    \item Transformer and Mamba-based methods (Section~\ref{sec:trans_mamba})
\end{itemize}

Additionally, Section~\ref{prelim} introduces fundamental concepts, benchmark datasets and evaluation metrics, while Section~\ref{discussion} identifies comparative analysis, current limitations and future research directions.



\section{Backgrounds and Preliminaries}
\label{prelim}
In this section, we present the fundamental concepts and background common to many LiDAR super-resolution methods. We detail data representations, problem formulation, evaluation metrics, benchmark datasets, and common challenges in the field.
\subsection{LiDAR Point Cloud Representation}
We focus on common automotive multi-channel LiDAR sensors, which capture 3D environments by rotating multiple laser beams and measuring distances via time-of-flight. A single emitted laser pulse can produce multiple echoes, resulting in multiple 3D points. Raw point cloud data can be represented in three main formats. First, spherical coordinates preserve the sensor format using the measured range and beam direction. Second, the 3D point cloud consists of unordered points in Euclidean space, accommodating all echo returns. Each point contains xyz coordinates and intensity values. Third, the range image representation projects 3D points onto a 2D image where pixel values represent depth information. The dimensions typically are 16, 32, 64, or 128 vertical channels (beams) by 1024 or 2048 horizontal resolution. Most SR methods adopt the range image representation due to its regular grid structure. However, this projection introduces quantization errors and potential information loss at object boundaries. Furthermore, multiple echoes along the same laser beam fall into the same grid cell and cannot be easily represented.

\begin{table*}[t]
\centering
\caption{Summary of LiDAR datasets for super-resolution research. It shows sensor configurations and annotation types.}
\label{dataset}
\begin{tblr}{
  vline{2} = {-}{},
  hline{1,13} = {-}{0.08em},
  hline{2} = {-}{},
}
Dataset       & LiDAR SR methods                                                                                     & Type      & LiDAR Sensors                                          & Channels                           & Annotation \\
KITTI raw \cite{kitti_raw}     & {\cite{yang2024tulip,goo2025flash,eskandar2022hals,triess2019cnn}}                                                                   & Real      & Velodyne HDL-64E                                       & 64                                 & -          \\
KITTI object \cite{kitti_raw}  & \cite{eskandar2022hals}                                                                                                 & Real      & Velodyne HDL-64E                                       & 64                                 & 3D Boxes   \\
SemanticKITTI \cite{behley2019semantickitti} & {\cite{gkillas2025guided,he2023lsr,chen2025srmamba,chen2025srmambav2}}                                                           & Real      & Velodyne HDL-64E                                       & 64                                 & Point-wise \\
SemanticPOSS \cite{pan2020semanticposs}  & \cite{gkillas2025guided}                                                                                               & Real      & PANDORA 40 Channel                                     & 40                                 & Point-wise \\
CARLA \cite{kwon2022implicit}         & \cite{yang2024tulip,kwon2022implicit,park2023implicit}                                                                                      & Synthetic & Synthetic LiDAR                                        & {16, 64, \\128, 256}               & 3D Boxes   \\
DurLAR \cite{li2021durlar}        & \cite{yang2024tulip}                                                                                                & Real      & Ouster OS1-128                                         & 128                                & 3D Boxes   \\
LiDAR-CS \cite{fang2024lidar}      & -                                                                                                    & Synthetic & {VLD-16, VLD-32, \\Velodyne HDL-64E, ONCE-40, \\Livox} & {16, 32, \\64, 40, \\None for Livox} & 3D Boxes   \\
LeGO-LOAM \cite{shan2018lego}     & \cite{tian2022lidar}                                                                              & Real      & Velodyne VLP-16                                        & 16                                 & -          \\
Ouster \cite{shan2020simulation}       & {\cite{tian2022lidar,gkillas2023efficient,shan2020simulation,xi2024fotv,gkillas2023federated,gkillas2023federated2,wang2023swin}} & Real      & Ouster OS1-64                                          & 64                                 & -          \\
nuScenes \cite{caesar2020nuscenes}     & \cite{eskandar2022hals,chen2025srmamba,chen2025srmambav2}                                                                             & Real      & {Velodyne HDL-32E}                                   & 32                                 & 3D Boxes   \\
Semantics \cite{semantics_dataset}     & \cite{triess2019cnn}                                                                                           & Real      & Velodyne VLP-32                                        & 32                                 & Point-wise 
\end{tblr}
\end{table*}

\subsection{Problem Formulation}
The LiDAR super-resolution task aims to reconstruct a high-resolution point cloud $\mathbf{P}_h \in \mathbb{R}^{n_h \times 3}$ from a low-resolution input $\mathbf{P}_l \in \mathbb{R}^{n_l \times 3}$ captured by a sensor with $H_l$ vertical channels, where $h$ and $l$ denote high and low resolution respectively, and $n_h$, $n_l$ represent the number of points in each point cloud data ($n_h>n_l$). The objective is to synthesize the output equivalent to a sensor with $H_h = \beta \times H_l$ channels, where $\beta$ represents the upsampling factor. Note that LiDAR super-resolution typically only targets vertical resolution enhancement by a factor of $\beta$, preserving the horizontal dimension $W$.

The low-resolution point cloud $\mathbf{P}_l$ is first converted to a range image $\mathbf{I}_l \in \mathbb{R}^{H_l \times W}$ through spherical projection, where each pixel stores the Euclidean distance $r = \sqrt{x^2 + y^2 + z^2}$ from the sensor origin. The horizontal dimension $W$ (typically 1024 or 2048) remains unchanged during upsampling. The projection mapping is defined as:
\begin{equation}
u = \frac{W}{2} - \frac{W}{2\pi} \cdot \arctan2(y, x)
\end{equation}
\begin{equation}
v = \frac{H}{\Theta_{\text{max}} - \Theta_{\text{min}}} \cdot \left(\Theta_{\text{max}} - \arctan\left(\frac{z}{\sqrt{x^2 + y^2}}\right)\right)
\end{equation}
where $(u, v)$ represent the image coordinates, and $\Theta_{\text{max}}$, $\Theta_{\text{min}}$ define the sensor's vertical field of view boundaries.

The SR network learns to predict a high-resolution range image $\mathbf{I}_h \in \mathbb{R}^{H_h \times W}$ from the low-resolution input, where $H_h = \beta \times H_l$.

The reconstruction is supervised using the L1 loss between predicted and ground truth range images:
\begin{equation}
\mathcal{L} = \|\mathbf{I}_h - \hat{\mathbf{I}}_h\|_1
\end{equation}
where $\mathbf{I}_h$ denotes the ground-truth high-resolution range image and $\hat{\mathbf{I}}_h$ represents the network prediction. However, some model-based approaches \cite{gkillas2023federated,gkillas2023federated2} have adopted L2 loss. Additionally, recent methods such as HALS \cite{eskandar2022hals} and SRMambaV2 \cite{chen2025srmambav2} combine domain-specific auxiliary losses with the L1 loss, including surface normal loss and Bird's-Eye-View consistency constraints. After obtaining the high-resolution range image, the image is reconstructed into a 3D point cloud through inverse spherical projection:
\begin{equation}
\begin{cases}
x = r \cdot \cos(\omega) \cdot \cos(\alpha) \\
y = r \cdot \cos(\omega) \cdot \sin(\alpha) \\
z = r \cdot \sin(\omega)
\end{cases}
\end{equation}
where $\alpha$ and $\omega$ are the azimuth and elevation angles derived from the pixel coordinates $(u, v)$. This back-projection step converts the super-resolved range image to a dense 3D point cloud. The entire process preserves the geometric structure of the LiDAR data while leveraging efficient 2D processing techniques. (See Figure~\ref{fig:intro})

\subsection{Evaluation Metrics}

Performance evaluation in LiDAR super-resolution requires both 2D and 3D assessment metrics to comprehensively measure reconstruction quality. For 2D range image evaluation, Mean Absolute Error (MAE) quantifies the pixel-wise reconstruction accuracy:
\begin{equation}
\text{MAE} = \frac{1}{N} \sum_{i=1}^{N} |r_i - \hat{r}_i|
\end{equation}
where $r_i$ and $\hat{r}_i$ represent the ground truth and predicted range values respectively, and $N$ is the total number of pixels. This metric directly measures how well the network predicts intermediate depth values between existing scan lines.

For 3D point cloud evaluation, Chamfer Distance (CD) serves as a primary metric, calculating the bidirectional average nearest-neighbor distance between predicted and ground truth point sets:
\begin{equation}
\begin{aligned}
\text{CD}(P_1, P_2)
&= \frac{1}{|P_1|} \sum_{p \in P_1} \min_{q \in P_2} \|p - q\|_2 \\
&\quad + \frac{1}{|P_2|} \sum_{q \in P_2} \min_{p \in P_1} \|q - p\|_2
\end{aligned}
\end{equation}
where $P_1$ and $P_2$ are the two point sets being compared. Specifically, $p$ and $q$ denote individual points belonging to the sets $P_1$ and $P_2$, respectively. This metric captures both completeness and accuracy by measuring distances from each point to its nearest neighbor in the opposite set.

Intersection over Union (IoU) provides a volumetric assessment crucial for understanding object reconstruction quality. The point clouds are first voxelized into a 3D grid with resolution $v$ (typically 0.1m), creating binary occupancy grids. Each voxel is marked as occupied if it contains at least one point. The IoU is then computed as:
\begin{equation}
\text{IoU} = \frac{|V_{\text{pred}} \cap V_{\text{gt}}|}{|V_{\text{pred}} \cup V_{\text{gt}}|}
\end{equation}
where $V_{\text{pred}}$ and $V_{\text{gt}}$ represent the sets of occupied voxels in predicted and ground truth clouds respectively. This metric is particularly valuable for autonomous driving applications as it directly relates to object detection performance. Related metrics include Precision ($\frac{|V_{\text{pred}} \cap V_{\text{gt}}|}{|V_{\text{pred}}|}$), measuring the fraction of predicted voxels that are correct, and Recall ($\frac{|V_{\text{pred}} \cap V_{\text{gt}}|}{|V_{\text{gt}}|}$), measuring the fraction of ground truth voxels that are recovered. The F1-score, computed as the harmonic mean of precision and recall:
\begin{equation}
\text{F1} = 2 \times \frac{\text{Precision} \times \text{Recall}}{\text{Precision} + \text{Recall}}
\end{equation}
provides a balanced measure of reconstruction accuracy. These metrics are essential for evaluating how well SR methods preserve object boundaries and avoid generating spurious points.
\subsection{Benchmark Datasets}
Table \ref{dataset} categorizes primary datasets into real-world and synthetic domains. Prominent real-world benchmarks include KITTI (64-channel) \cite{kitti_raw,behley2019semantickitti}, nuScenes (32-channel) \cite{caesar2020nuscenes}, and DurLAR (128-channel) \cite{li2021durlar}. Due to the infeasibility of capturing perfectly aligned paired data in real-world scenarios, low-resolution inputs were generated by downsampling these high-resolution captures.

Synthetic datasets overcome alignment limitations by simulating multi-resolution sensors in virtual environments. The CARLA dataset \cite{kwon2022implicit, dosovitskiy2017carla} generates perfectly aligned point clouds across varying channel counts (16 to 256), eliminating calibration errors. To address cross-sensor domain gaps, LiDAR-CS \cite{fang2024lidar} utilizes a pattern-aware simulator to generate data from six distinct sensor types (Velodyne variants from 16 to 128 channels \cite{ouster2024}, ONCE-40 \cite{mao2021one}, and Livox sensors \cite{livox2024}) for identical scenarios which enables the study of sensor-specific characteristics.

\section{Convolutional Neural Network-based methods}
\label{sec:CNN}
Convolutional Neural Networks (CNNs) were among the first deep learning architectures applied to LiDAR super-resolution. This section examines the progression from early CNN designs that were built for image processing to advanced models tailored for the unique characteristics of LiDAR data.

\subsection{Early CNN architectures}
CNN-based LiDAR super-resolution began by adapting image SR ideas to 3D point clouds. A range image is used to represent the scan, and the model upsamples only the vertical resolution \cite{chen2021channel}. The network follows a UNet-style design with a Channel Attention-based Reconstruction Block that learns channel importance. Circular padding handles wraparound at borders and helps edge recovery on range images \cite{chen2021channel}. In addition, a two–stream fusion of \emph{xyz} and range features is used before reconstruction, which further stabilizes edge regions \cite{chen2021channel}.

To address the over-smoothing artifacts common in pixel-wise regression, a CNN-based approach prioritizing perceptual realism is proposed \cite{triess2019cnn}. Instead of relying solely on distance metrics, this method utilizes a perceptual loss derived from a frozen semantic segmentation network ($\phi$) to enforce high-level feature consistency:
\begin{equation}
\mathcal{L}_{\text{feat}}=\sum_{c,i,j}\left|\phi\!\left(r^{gt}\right)_{cij}-\phi\!\left(r^{pred}\right)_{cij}\right|
\end{equation}
Furthermore, a semantic consistency term is employed to keep class boundaries sharp by minimizing the cross-entropy loss between the semantic labels of the predicted and ground truth range images \cite{triess2019cnn}.

\subsection{Advanced CNN architectures}
HALS \cite{eskandar2022hals} observes that range statistics vary with height. Upper beams often have larger mean distance and variance. HALS uses two upsampling branches with different receptive fields and each branch outputs a confidence map to model uncertainty. The final prediction blends the branches with softmax confidences. The method regresses polar coordinates to reduce vertical quantization error and adds a surface normal loss to preserve structure.

LiDAR-SR \cite{shan2020simulation}, a simulation-based approach, trains only on CARLA and adapts to real scenes. By employing Monte Carlo dropout to approximate a Bayesian Neural Network, the model estimates predictive uncertainty and filters out noisy points based on their variance. The authors report improved occupancy mapping quality when uncertain pixels are filtered before fusion.

\subsection{Alternative Methods}
LSR-RIBNet \cite{he2023lsr} fuses range and intensity and azimuth (bearing–angle) images. Dense connections extract rich features and a hybrid attention module applies channel attention followed by spatial attention. The azimuth image helps preserve edges and the model remains lightweight while improving PSNR and MAE.

One study proposes a two-stage pipeline for pedestrian orientation \cite{lee2023pedestrian}. It applies ESRGAN \cite{wang2018esrgan} to super-resolve LiDAR range, intensity, and near-infrared images. And then input the results to a ResNet \cite{he2016deep} classifier to predict eight discrete orientation bins. The super-resolved inputs results in higher orientation accuracy than the original low resolution images.

\section{Model-Based Deep Unrolling and Federated Methods}
\label{sec:DU_FM}
This section reviews two categories of LiDAR super-resolution that emphasize both computational efficiency and model interpretability: model-based deep unrolling and its extension to federated learning. While purely data-driven approaches, such as CNNs, have achieved significant performance, they often operate as ‘black-box’ models requiring massive parameters to approximate the mapping function. To address the lack of interpretability and parameter efficiency, Model-Based Deep Unrolling (DU) integrates physical degradation models directly into the network architecture. Both start from the same degradation model,
\begin{equation}
Y = SX + N,
\end{equation}
where $X$ is the high-resolution range image, $S$ is a known downsampling operator, $Y$ is the low-resolution observation, and $N$ is noise. The solution alternates two parts: a data-consistency step that follows from the model, and a learned regularizer (a denoiser). In practice, the data step often uses a closed-form update, for example
\begin{equation}
X \leftarrow (S^\top S + \beta I)^{-1}(S^\top Y + \beta Z),
\end{equation}
and the regularizer sets $Z \leftarrow f_\theta(X)$, where $f_\theta$ is a small CNN. Deep unrolling builds a $K$-layer network by unrolling $K$ iterations of this scheme. This yields a compact, stable model by constraining the search space and limiting learning to the regularizer.

\subsection{Model-Based Deep Unrolling Networks}
An efficient unrolled SR model\cite{gkillas2023efficient,gkillas2023federated} uses a small CNN as the regularizer inside a $K$-step Half-Quadratic Splitting (HQS)/Alternating Direction Method of Multipliers (ADMM)-like scheme. With about $0.1$M parameters, the method\cite{gkillas2023efficient} cuts the parameter count by about $99.75\%$ compared to heavy baselines such as LiDAR-SR \cite{shan2020simulation} and still improves L1 error. It also helps downstream SLAM. When the high-resolution output is projected back to 3D and passed to LeGO-LOAM, the absolute trajectory error is lower than with the low-resolution input. The gain comes from the model-based data step, which gives a strong prior on the solution, and from the fact that the network learns only the denoiser. 
Because the closed-form update already enforces the image formation model, a shallow denoiser with four or five layers is enough, which also makes inference fast and communication light.

FOTV-HQS \cite{xi2024fotv} replaces the standard total variation prior with a fractional-order TV (Total Variation) prior. The network alternates between the data term and the FOTV (Fractional-Order Total Variation) term, each implemented by a light CNN module within HQS. This improves edge and texture preservation at $4\times$, $8\times$, and $16\times$ upscaling, and reports better PSNR and SSIM with large parameter savings (about $99.68\%$ lower than LiDAR-SR \cite{shan2020simulation}) . The idea is that standard TV can cause staircasing, while fractional-order gradients keep high-frequency content. Training uses paired range images such as $64 \times 1024$ and $16 \times 1024$, with simulation for training and real scans for testing. A simple objective is
\begin{equation}
\min_X \tfrac{1}{2}\|Y - SX\|_F^2 + \lambda\,\mathrm{TV}^\alpha(X),
\end{equation}
where $\alpha$ represents the fractional order and $\lambda$ is the regularization parameter \cite{xi2024fotv}.

Guided model-based SR \cite{gkillas2025guided} integrates the unrolled SR module with a semantic segmentation backbone and trains them jointly. A learnable mask gives higher weight to important regions and small classes such as bicycle and person. This joint setup improves mIoU compared to training SR and segmentation in isolation. The model runs at about 23\,fps and uses about $1\%$ of the parameters of transformer-based SR, yet it approaches the segmentation quality of a 64-beam input while using low-cost inputs and only a few unrolled steps. The approach is compatible with different range-view segmentation backbones and is easy to plug into existing pipelines.

\subsection{Federated Methods}
In standard federated learning, each client (for example, a vehicle) trains locally, and a server aggregates the model weights. Raw data stays on the client. Results show that federated training reaches L1 error close to central training and is clearly better than training each client alone \cite{gkillas2023federated2}. Federated methods prioritize privacy and reduced data movement but require tuning communication budgets to minimize the slight performance gap against central training.

Federated deep unrolling (FL-DU) \cite{gkillas2023federated} minimizes communication by exchanging only the learned denoiser. This interpretable framework matches centralized performance even under homomorphic encryption and outperforms autoencoder-based federated methods with faster convergence and 99.75\% fewer parameters \cite{gkillas2023efficient}.

\section{Implicit Representation and Continuous Learning Networks}
\label{sec:implicit}

Although model-based deep unrolling offers parameter efficiency, it remains constrained by fixed upsampling factors defined during training. To overcome this limitation, implicit representation has emerged as a transformative paradigm. Unlike traditional explicit methods that learn fixed resolution mappings, implicit functions are defined in continuous space, enabling resolution-free upsampling to arbitrary densities. This capability allows a single model to flexibly adapt to diverse sensor configurations. This section focuses on two key implicit methodologies: ILN \cite{kwon2022implicit} and IPF \cite{park2023implicit}.

\subsection{Implicit LiDAR Network (ILN)}

ILN \cite{kwon2022implicit} was the first practical implicit network designed specifically for LiDAR super-resolution. Unlike previous approaches that predict depth values directly, ILN introduces an innovative method of learning interpolation weights. Given the frequent occurrence of sharp depth changes in LiDAR data such as building edges and vehicle boundaries, ILN adopts the following formulation:

\begin{equation}
\hat{r} = \sum_{t=1}^{4} g(z'_t|\theta) \cdot r_t
\end{equation}

where $g(z'_t|\theta)$ represents the interpolation weight predicted by the network and $r_t$ denotes the actual measured depth value from neighboring pixels. The key advantage of this approach lies in the network learning how to blend existing measurements rather than generating new depth values. This ensures stable convergence even in early training stages since outputs remain close to input data.

Another significant contribution of ILN is the incorporation of the Transformer self-attention mechanism \cite{ashish2017attention,dosovitskiy2021an} to capture correlations between neighboring pixels. Each neighbor feature contains relative distance information with respect to the query position. Through self-attention, the network adaptively determines which neighbors are more important for depth prediction at the current query location. This nonlinear interpolation approach enables sharp reconstruction of boundary regions that linear interpolation would otherwise blur.

\subsection{Implicit Point Function (IPF)}

IPF \cite{park2023implicit} advances beyond the 2D depth map approach of ILN by proposing a true 3D implicit function that directly utilizes the geometric properties of point clouds. While ILN projects 3D points onto 2D depth maps before processing, IPF operates directly in 3D space to minimize geometric information loss.

The core idea of IPF involves moving neighboring points directly onto the query ray to generate target points when given a query ray:

\begin{equation}
\hat{p} = \sum_{t=1}^{4} w^*_t \cdot (p_{proj}^t + \delta_t \cdot r)
\end{equation}

where $p_{proj}^t$ is the orthogonal projection of the neighboring point onto the ray, $\delta_t$ is the learned depth offset, and $r$ is the query ray direction. Here, $w^*_t$ represents the aggregation weights predicted via an attention mechanism, determining the contribution of each neighbor. The innovation of this approach lies in constraining predicted points to lie precisely on the 3D query ray. Unlike ILN which is limited to weighted averages of neighbor values, IPF can predict values beyond the measurement range through the offset $\delta_t$, providing greater flexibility in handling abrupt depth changes.

\subsubsection{On-the-Ray Positional Encoding}

One of the most important technical contributions of IPF is the on-the-ray positional encoding. This method explicitly encodes the 3D geometric relationship between query rays and neighboring points using three vector components:

The \textbf{query ray direction} provides directional guidance for SR. The \textbf{projection vector} indicates the approximate position of neighboring points on the ray. The \textbf{rejection vector} captures the perpendicular distance between the ray and neighboring points.

By transforming these three vectors through high-frequency positional encoding \cite{mildenhall2021nerf} and concatenating them, IPF simultaneously captures local geometric relationships and global spatial context. This plays a crucial role in preserving fine 3D structural features that 2D-based methods might overlook.

\begin{table*}[h]
\centering
\caption{Comparative analysis of deep learning-based LiDAR super-resolution architectures. We explicitly contrast the key mechanisms, advantages, and limitations of each category}
\label{tab:comparison}
\resizebox{\textwidth}{!}{
\begin{tblr}{
  vline{2} = {-}{},
  hline{1-2,6} = {-}{},
}
Category                   & Representative Methods & Key Mechanisms                                                                                     & Pros                                                                                                               & Cons                                                                                                                                         \\
CNN-Based Models                & \cite{shan2020simulation,eskandar2022hals,he2023lsr,triess2019cnn,lee2023pedestrian}                   & {• Projects 3D points to 2D range images\\ • Uses Standard Convolutions or Polar coordinates.}     & {• Fast real-time inference\\ • Easy implementation\\ • Simple multi-modal fusion}                                 & {• Over-smoothing of object edges\\ • Fixed input/output resolution\\ • Performance drops at long range\\ • Lack of global context modeling} \\
{Model-Based \\Deep Unrolling} & \cite{gkillas2023federated,xi2024fotv,gkillas2025guided}                    & {• Unrolls optimization algorithms into network layers.\\ • Learns regularizers/denoisers.}        & {• Interpretable architecture.\\ • 99\% fewer parameters (lightweight).\\ • Enables Federated Learning (privacy).} & {• Limited non-linear expressiveness.\\ • Iterative steps may slow down inference.}                                                          \\
Implicit Representation    & \cite{kwon2022implicit,park2023implicit}                      & {• Learns continuous functions for resolution-free upsampling\\ • Projects to 3D query rays (IPF)} & {• Resolution-agnostic\\ • Preserves 3D geometry (IPF).\\ • Robust weight prediction (ILN).}                       & {• High computation cost for dense queries\\ • Potential information loss in 2D projection (ILN)}                                            \\
{Transformer \& Mamba\\-Based Models}        & \cite{yang2024tulip,goo2025flash,chen2025srmamba,chen2025srmambav2,wang2023swin}                  & {• Self-Attention\\ • State Space Models}                                                          & {• Captures long-range dependencies\\ • SOTA accuracy metrics\\ • Reduced artifacts}                               & {• High computational cost (Transformers)\\ • Higher latency than CNNs}                                                                      
\end{tblr}
}
\end{table*}

\subsection{Continuous Learning and Practical Implications}

Both ILN \cite{kwon2022implicit} and IPF \cite{park2023implicit} exhibit resolution-free characteristics that allow a single model to generate outputs at various resolutions. Even when trained only at $128 \times 2048$ resolution, these models can perform inference from $64 \times 1024$ to $256 \times 4096$. This capability enables adaptive resolution adjustment based on situational requirements in actual autonomous driving systems.


Experimental results demonstrate that while ILN offers fast convergence and stability through interpolation weight prediction, IPF achieves superior geometric fidelity by directly utilizing 3D spatial encoding and constraints. This evolution marks a paradigm shift from 2D-inspired approach (ILN) to 3D-native understanding (IPF), significantly enhancing robustness in challenging regions like occluded boundaries and ground points essential for autonomous driving.

\section{Transformer and Mamba-based Methods}
\label{sec:trans_mamba}
To address the limited receptive fields of implicit methods, Transformer and Mamba-based architectures leverage transformer \cite{dosovitskiy2021an} and state-space models \cite{gu2024mamba,zhu2024vision} to effectively capture global context on projected 2D range images \cite{chen2025srmamba,yang2024tulip}. These approaches typically target 4x vertical upsampling which utilize techniques like circular padding to preserve 360-degree field-of-view consistency while recovering fine details in sparse regions.

\subsection{Transformer-based Approaches}
TULIP\cite{yang2024tulip} pioneered the use of Swin-UNet \cite{cao2022swin} architecture for range image processing. The method divides each row into 1×4 patches which maintains vertical resolution. To handle the cylindrical nature of LiDAR data, TULIP uses rectangular attention windows with circular padding at image boundaries. This ensures smooth transitions at the 360-degree wraparound point. The reconstruction stage employs pixel shuffle \cite{shi2016real} for upsampling followed by Monte Carlo dropout \cite{gal2016dropout} filter. This filter runs multiple forward passes and removes predictions with high variance, reducing floating artifacts near object edges.

FLASH \cite{goo2025flash} extends TULIP design by processing features in both spatial and frequency domains. The method extracts frequency information by applying Fast Fourier transform (FFT) to channel-averaged features which helps preserve sharp boundaries and fine textures. For skip connections, the authors introduce an adaptive fusion mechanism that learns to select from multiple kernel scales at each position. These fused features then pass through lightweight attention modules for further refinement. This dual-domain approach improves multiple metrics including IoU, F1-score, and Chamfer Distance. FLASH achieves these gains without the computational overhead of Monte Carlo dropout.

Swin-T-NFC CRFs \cite{wang2023swin} integrates Swin transformers \cite{liu2021swin} with Conditional Random Fields \cite{lafferty2001conditional}. Unlike direct super-resolution methods, it refines a Gaussian-smoothed vertically upsampled range image. The CRF decoder uses windowed attention to compute per-pixel costs and pairwise potentials \cite{wang2023swin}, effectively balancing sharp boundary preservation with surface smoothness.
\subsection{State-Space Model Approaches}
SRMamba \cite{chen2025srmamba} introduces visual state-space models as an alternative to transformers which offer linear complexity instead of quadratic scaling with sequence length. SRMamba applies two preprocessing steps specific to LiDAR data. First, it uses Hough transforms to align horizontal scan lines to correct for sensor motion \cite{chen2025srmamba}. Second, it fills linear gaps along rows through interpolation to address the structured sparsity patterns in range images. The architecture follows a U-Net \cite{ronneberger2015u} design with state-space blocks. Combined with pixel shuffle upsampling, this approach shows consistent improvements across both indoor and outdoor datasets \cite{chen2025srmamba}.

SRMambaV2 \cite{chen2025srmambav2} refines the state-space approach through a three-stage pipeline called: scan, modulate, and focus. The scan stage uses state-space blocks to gather global context across the entire range image. The modulate stage introduces a dual-branch design: one branch expands the receptive field for broader context while another preserves fine details through channel-wise modulation. The focus stage then applies Swin transformer \cite{liu2021swin} blocks to capture local structures and edges. For training, SRMambaV2 uses a composite loss that combines standard L1 reconstruction with a learnable importance mask and a bird's-eye-view consistency term \cite{chen2025srmambav2}.

\section{Discussion and Future Directions}
\label{discussion}
\subsection{Comparative Analysis}
To provide a clear conceptual framework, Table \ref{tab:comparison} summarizes the key trade-offs of the four categories. \textbf{Model-Based Deep Unrolling (DU)} excels in interpretability and parameter efficiency, making it uniquely suited for bandwidth-constrained scenarios like Federated Learning \cite{gkillas2023federated, xi2024fotv}. However, reliance on fixed physical degradation models may limit its expressiveness for complex semantic features compared to purely data-driven approaches. \textbf{Implicit Representation} methods overcome fixed-resolution constraints through continuous function learning, offering resolution-agnostic upsampling critical for cross-sensor compatibility \cite{kwon2022implicit, park2023implicit}. The primary drawback is the high computational cost required to query dense points during inference. \textbf{CNN-based architectures} serve as a robust baseline, leveraging mature optimizations to ensure fast, real-time inference \cite{shan2020simulation, he2023lsr}. Yet, due to limited receptive fields, they often fail to capture global context, resulting in over-smoothed object boundaries. \textbf{Transformer and Mamba-based methods} represent the current state-of-the-art, effectively capturing long-range dependencies for superior geometric consistency \cite{yang2024tulip}. While standard attention mechanisms incur high computational costs, recent innovations in frequency domain processing \cite{goo2025flash} and state-space models \cite{chen2025srmamba, chen2025srmambav2} have successfully mitigated latency issues to achieve real-time performance.

\subsection{Current Limitations}
Despite the substantial progress in LiDAR super-resolution, critical challenges remain for practical deployment. \textbf{Cross-sensor generalization} is a major limitation. For example, models trained on Velodyne sensors often fail on Livox or different LiDAR products due to different beam patterns and noise characteristics \cite{fang2024lidar}. Due to this limitation, the models are forced to retrain for each sensor type. Another challenge is achieving \textbf{real-time inference}: autonomous driving requires over 25 fps processing \cite{maksymova2018review,dai2022requirements}, yet despite a 99\% parameter reduction via model-based unrolling, a performance gap remains for embedded systems. Furthermore, most existing work \cite{goo2025flash, shan2020simulation, chen2025srmamba, chen2025srmambav2} focuses on reconstruction metrics rather than \textbf{downstream task performance}. The lack of systematic evaluation on object detection or segmentation leaves it unverified whether SR effectively bridges the sensor gap beyond mere visual improvement. Moreover, dominant range-view methods inherently suffer from geometric information loss due to spherical projection. While \textbf{3D-native (point-based) approaches} could resolve this, they are currently impractical for autonomous driving due to high computational costs in large-scale outdoor scenes \cite{chen2025srmamba}.

\subsection{Future Research}
Several directions can be proposed to advance this field. \textbf{Hybrid domain processing} shows strong potential, with methods such as FLASH \cite{goo2025flash} which demonstrated that combining frequency and spatial domains improves boundary preservation. \textbf{Combining neural implicit functions with grid-based explicit features} could benefit from both continuous geometry modeling and efficient discrete computation. \textbf{Self-supervised learning} represents another option as obtaining paired high and low-resolution data is difficult in practice. \textbf{Multi-modal fusion with camera or intensity data} could provide a form of guidance for SR. RGB images or intensity data offer dense texture information to resolve ambiguities in sparse point clouds. 
Despite the potential of \textbf{foundation models pre-trained on diverse LiDAR datasets} for zero-shot super-resolution across sensors, the majority of current methods still formulate the problem through 2D projection-based representations \cite{LVM_lidar_goo,thengane2025foundational}.

\section{Conclusion}
This paper presented the first comprehensive survey of LiDAR super-resolution for autonomous driving. We reviewed the evolution from CNN-based baselines to Model-based Deep Unrolling.
Furthermore, we analyzed implicit representations for resolution flexibility and Transformer/Mamba-based approaches for capturing global context. Critical challenges in cross-sensor generalization persist. Future research should prioritize sensor-agnostic architectures via self-supervised and multi-modal fusion. Furthermore, LiDAR super-resolution is vital to bridge the gap between high-end and low-cost sensors for accessible autonomous driving.


\bibliographystyle{IEEEtran}
\bibliography{IEEEabrv,mybibfile}







\end{document}